\documentclass[a4paper,fleqn]{cas-dc}

\usepackage{booktabs,caption}
\usepackage[flushleft]{threeparttable}

\usepackage[authoryear]{natbib}
\usepackage{algorithm}
\usepackage{algpseudocode}
\usepackage{xcolor}
\usepackage{array}

\usepackage{caption}

\def\tsc#1{\csdef{#1}{\textsc{\lowercase{#1}}\xspace}}
\tsc{WGM}
\tsc{QE}

\begin{document}
\let\WriteBookmarks\relax
\def\floatpagepagefraction{1}
\def\textpagefraction{.001}

\shorttitle{}

\title [mode = title]{Leveraging Optimal Transport for Enhanced Offline Reinforcement Learning in Surgical Robotic Environments}  

\tnotemark[<tnote number>] 


%

\author[1]{Maryam Zare}[]

\cormark[1]


\ead{mzare@deakin.edu.au}



\affiliation[1]{organization={Institute for Intelligent Systems Research and Innovation (IISRI) Deakin University},
            addressline={Waurn Ponds}, 
            city={},
          citysep={}, 
            postcode={3216}, 
            state={VIC},
            country={Australia}}

\author[1]{Parham M. Kebria}[]
\author[1]{Abbas Khosravi}[]






\cortext[1]{Corresponding author}



\begin{abstract}
Most Reinforcement Learning (RL) methods are traditionally studied in an active learning setting, where agents directly interact with their environments, observe action outcomes, and learn through trial and error. However, allowing partially trained agents to interact with real physical systems poses significant challenges, including high costs, safety risks, and the need for constant supervision. Offline RL addresses these cost and safety concerns by leveraging existing datasets and reducing the need for resource-intensive real-time interactions. Nevertheless, a substantial challenge lies in the demand for these datasets to be meticulously annotated with rewards. In this paper, we introduce Optimal Transport Reward (OTR) labelling, an innovative algorithm designed to assign rewards to offline trajectories, using a small number of high-quality expert demonstrations. The core principle of OTR involves employing Optimal Transport (OT) to calculate an optimal alignment between an unlabeled trajectory from the dataset and an expert demonstration. This alignment yields a similarity measure that is effectively interpreted as a reward signal. An offline RL algorithm can then utilize these reward signals to learn a policy. This approach circumvents the need for handcrafted rewards, unlocking the potential to harness vast datasets for policy learning. Leveraging the SurRoL simulation platform tailored for surgical robot learning, we generate datasets and employ them to train policies using the OTR algorithm. By demonstrating the efficacy of OTR in a different domain, we emphasize its versatility and its potential to expedite RL deployment across a wide range of fields.
\end{abstract}




\maketitle

\section{Introduction}
In recent years, significant advancements have transformed the landscape of RL, with offline RL emerging as a transformative framework \citep{levine2020offline}. This paradigm offers the potential to derive optimal decision-making policies from existing datasets, bypassing the need for real-time interaction with the environment. Recently, many offline RL algorithms have been proposed for learning from previously collected diverse and suboptimal datasets \citep{kumar2020conservative, luo2023optimal, zhou2023real, wang2020critic}. This is particularly valuable in contexts where data collection can be resource-intensive or where safety constraints limit direct interaction, such as in surgical robotics \citep{WANG2022103945}. Offline RL enables policy improvement beyond the limitations of the initial data distribution, making it a compelling avenue for domains characterized by complex, high-dimensional state spaces.

However, the success of offline RL hinges on the availability of well-defined reward functions \citep{yu2022leverage}. These reward signals play a crucial role in guiding the learning process by providing a measure of the desirability of different states and actions. Unfortunately, specifying reward functions can be a daunting task, especially in intricate domains like surgical robotics, where the criteria for success are often multifaceted and nuanced \citep{singh2019end}. Traditional approaches to reward engineering can be time-consuming, labor-intensive, and may not fully capture the intricacies of expert behavior.

Navigating the complexities of intricate domains, particularly in the absence of clear reward functions, demands innovative approaches. One powerful way to address this challenge involves tapping into expert demonstrations—an established avenue across diverse fields. These demonstrations provide a direct means of accessing valuable insights from experts without the need for labor-intensive reward engineering. An example of harnessing expert demonstrations is found in Imitation Learning (IL), a well-established framework. IL focuses on mimicking expert actions to learn policies, as demonstrated by works such as \citep{schaal1999imitation, zare2023survey, osa2018algorithmic, abbeel2010autonomous, dadashi2020primal}. IL's strength lies in its capacity to learn expert behaviors without imposing the requirement for explicit reward specifications \citep{TSURUMINE2022104264}.

\begin{figure*}[!t]
\centering
\includegraphics[width = .99\linewidth]{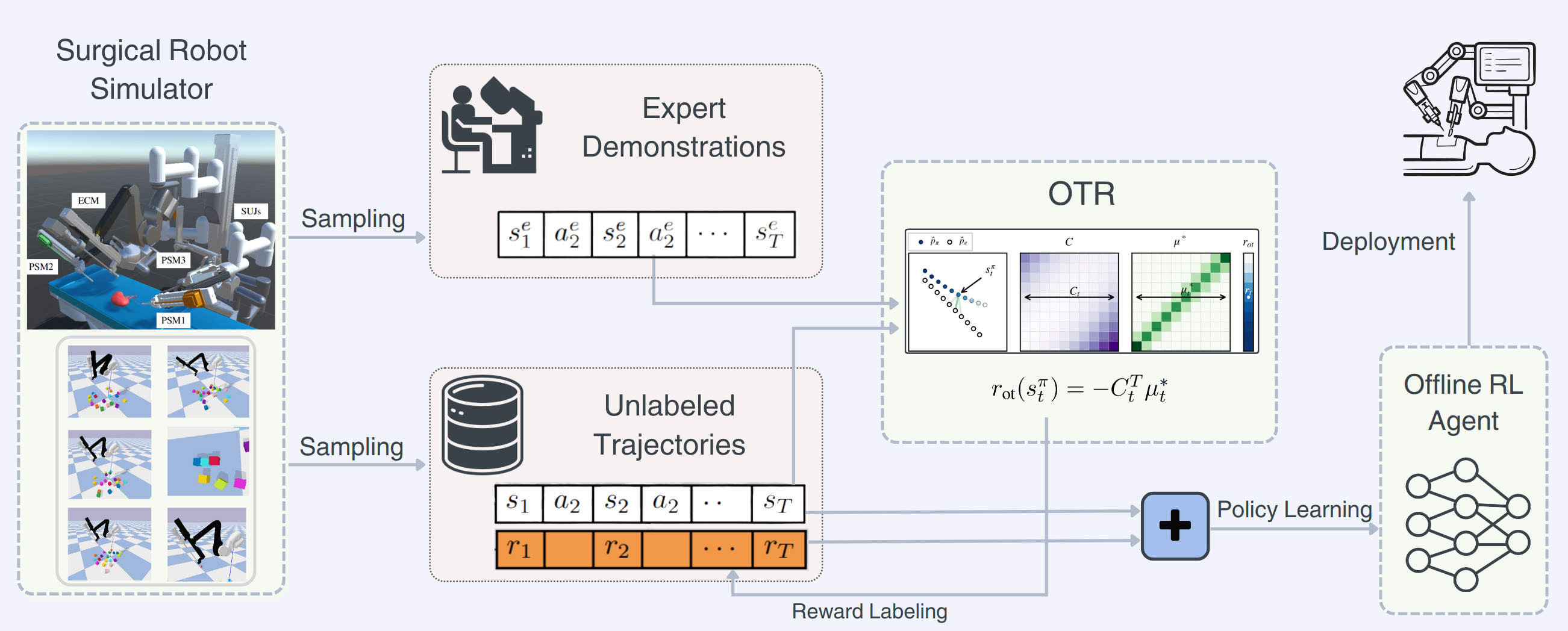}

\caption{\textbf{Overview of the Proposed Framework.}This figure illustrates the sequential stages of the proposed methodology. Initially, expert demonstrations and unlabeled trajectories are collected through the SurRol simulation platform. Subsequently, leveraging expert demonstrations and the offline dataset that lacks reward labels, OTR , effectively augments the offline dataset by integrating reward labels, denoted as $r_i$, using OT techniques \citep{luo2023optimal}. The generated labeled dataset, rich in reward information, becomes a valuable resource for training an offline RL algorithm. Finally, the learned policy is deployed and evaluated in the simulation environment.}\label{pipeline}
\end{figure*}

Building upon the foundation of expert-guided policy learning emerges an innovative and promising approach—the integration of OT theory into offline RL. OT \citep{villani2009optimal, peyre2019computational}, a mathematical framework proficient in measuring distances between probability distributions, offers a structured approach for aligning and comparing trajectories. Leveraging OT, the OTR algorithm \citep{luo2023optimal} introduces an elegant strategy that circumvents the need for manual reward assignment. By aligning trajectories from unlabeled data with expert demonstrations, OTR automatically assigns reward values. This alignment transforms the dataset into one equipped with expert-informed reward information, setting the stage for offline RL algorithms to learn optimal policies that mimic expert behavior. Figure \ref{pipeline} illustrates the overall framework of this paper. The fusion of expert demonstrations and OT-based reward assignments, as exemplified by OTR, showcases the innovative combinations that can broaden the horizons of offline RL, offering fresh avenues for policy learning in challenging settings.

Within the scope of this paper, we have undertaken a comprehensive exploration of the OTR algorithm, assessing its effectiveness in the context of surgical robotics. We employ the SurRoL simulation platform \citep{xu2021surrol} as the foundation for our evaluation, utilizing its capability to generate datasets that authentically capture the intricacies of surgical procedures. This platform serves as a controlled yet immersive environment to rigorously test and validate the proposed approach. The integration of the OTR algorithm within this simulated surgical context holds a significance that extends beyond the realm of autonomous surgical execution. It presents the potential to transform various other domains characterized by intricate decision-making tasks under constrained settings, marking a significant stride towards adaptable solutions in a wide array of real-world applications.

\section{Preliminaries}
In this section, we introduce the concept of OT, which forms the basis for OTR approach to labeling previously unlabeled datasets.

We establish the learning framework within an episodic, finite-horizon Markov Decision Process (MDP) denoted as \((\mathcal{S}, \mathcal{A}, p_0(s), p(s_0|s, a), r(s, a), \gamma)\), where \(\mathcal{S}\) denotes the state space, \(\mathcal{A}\) signifies the action space, \(p_0(s)\) represents a distribution of initial states, \(p(s_0|s, a)\) characterizes the environment dynamics, \(r(s, a)\) encapsulates the reward function, and \(\gamma\) corresponds to the discount factor. The agent interacts with the MDP following a policy \(\pi(a|s)\). The agent's interactions with the environment generate state-action trajectories denoted as \(\tau = (s_1, a_1, s_2, a_2, \ldots, s_T)\), where $T$ represents the episode horizon.

\subsection{Offline reinforcement learning}
In the context of RL, offline RL stands as a unique paradigm wherein an agent endeavors to learn an effective policy from an \textit{static} dataset of trajectories $\mathcal{D} = {(s_i^t, a_i^t, s_i^{t+1}, r_i^t)}$. The primary objective revolves around acquiring a policy that maximizes the cumulative discounted rewards. The main distinction between offline RL and traditional RL lies in the absence of online interactions with the environment. Essentially, offline RL requires the learning algorithm to be able to learn enough information about the dynamics of the MDP by using a fixed dataset and then to learn a policy $\pi$ that achieves the maximum cumulative reward when it is actually used to interact with the MDP \citep{levine2020offline}.

\subsection{Understanding optimal transport}
OT is a mathematical framework that finds the optimal way to transfer mass from one distribution to another while minimizing transportation costs. This concept has found applications in a range of fields, from economics to computer vision \citep{peyre2019computational}. In our context, OT serves as a powerful tool for aligning trajectories between different distributions, a key process in assigning reward values to unlabeled data within the framework of offline RL.

\begin{algorithm*}[!t]
\caption{Optimal Transport Reward}
\label{alg:otr}
\begin{algorithmic}[1]
\Function{OptimalTransportReward}{$\text{expert\_demonstrations}, \text{unlabeled\_trajectories}$}
    \State $\text{labeled\_trajectories} \gets []$
    \For{$\text{agent\_trajectory}$ in $\text{unlabeled\_trajectories}$}
        \State $\text{cost\_matrix, optimal\_coupling\_matrix} \gets \text{SolveOT}(\text{expert\_demonstrations}, \text{agent\_trajectory})$ 
        \State $\text{rewards} \gets \text{ComputeRewards}(\text{cost\_matrix, coupling\_matrix})$
        \State $\text{labeled\_trajectory} \gets \text{agent\_trajectory}$ with added $\text{rewards}$
        \State $\text{labeled\_trajectories.append(labeled\_trajectory)}$
    \EndFor
    \State \Return $\text{labeled\_trajectories}$
\EndFunction
\end{algorithmic}
\end{algorithm*}

Consider two distributions, represented by probability measures $\mathcal{P}$ and $\mathcal{Q}$, each defined over a finite set $\mathcal{X}$. These distributions are characterized by probability mass functions $p(x)$ and $q(y)$, respectively. The cost associated with transporting unit mass from element $x$ to element $y$ is governed by the cost function $c(x,y)$. The core objective of OT is to determine a transportation plan $\mu$ that minimizes the aggregate transportation cost while ensuring mass conservation across distributions:

\begin{equation}
\sum_{y \in \mathcal{X}} \mu(x,y) = p(x), \quad \sum_{x \in \mathcal{X}} \mu(x,y) = q(y) 
\end{equation}

Here, $\mu(x,y)$ denotes the amount of mass transported from $x$ to $y$. The aggregate cost of transporting mass according to the plan $\mu$ is given by:

\begin{equation}
C(\mu) = \sum_{x,y \in \mathcal{X}} c(x,y) \cdot \mu(x,y) 
\end{equation}

The (squared) Wasserstein distance, serving as a metric to gauge the divergence between distributions, is characterized as the minimum achievable cost of transporting mass and is defined as:

\begin{equation}
W^2(\mathcal{P}, \mathcal{Q}) = \min_{\mu \in \Gamma(\mathcal{P}, \mathcal{Q})}C(\mu) 
\end{equation}

Here, $\Gamma(\mathcal{P}, \mathcal{Q})$ denotes the set of all possible transportation plans that align distribution $\mathcal{P}$ with distribution $\mathcal{Q}$ while preserving mass conservation constraints.

In the context of offline RL, we extend the principles of OT to align trajectories from expert demonstrations and unlabeled trajectories. Let $\hat{p}^\text{e}$ and $\hat{p}^\pi$ represent the empirical state distribution of expert and unlabeled trajectories, computed as $\frac{1}{T'} \sum_{t=1}^{T'} \delta_{s^\text{e}_t}$ and $\frac{1}{T} \sum_{t=1}^{T} \delta_{s^\pi_t}$ respectively. Here, $\delta_{s}$ represent the Dirac measures for $s$. Utilizing these notations, the squared Wasserstein distance between $\hat{p}^\pi$ and $\hat{p}^\text{e}$, denoted as $W^2(\hat{p}^\pi, \hat{p}^\text{e})$

\begin{equation}
W^2(\hat{p}_\pi, \hat{p}_e) = \min_{\mu \in \Gamma} \sum_{t=1}^{T} \sum_{t'=1}^{T'} c(s_\pi^t, s_e^{t'})\mu_{t,t'}
\end{equation}

\noindent quantifies the divergence between the distributions of states in the expert and unlabeled trajectories.

OTR leverages the foundational principles of OT to introduce an innovative labeling technique. This technique revolves around aligning distributions of states, allowing it to autonomously assign reward values to previously unlabeled trajectories. The key components of this process include the utilization of a cost matrix $C_{t,t'} = c(s_\pi^t, s_e^{t'})$ and the identification of an optimal transportation plan $\mu^*_{t,t'}$ that minimizes the Wasserstein distance. The rewards can then be computed as follow:

\begin{equation}
r_\text{OT}(s_t^\pi) = - \sum_{t'=0}^{T'} c(s_t^\pi, s^\text{e}_{t'}) \mu^{*}_{t, t'}  \label{eq:reward_computation}
\end{equation}

This reward formulation guides the policy towards imitating expert trajectories, emphasizing the integral role of OT in seamlessly melding expert behavior with policy optimization.

\subsection{Leveraging optimal transport in offline reinforcement learning}

Having comprehensively explored the foundational principles of OT, we now move on to its application in the context of offline RL. Algorithm \ref{alg:otr} presents OTR pseudocode, which systematically aligns expert demonstrations with unlabeled trajectories and assigns rewards accordingly. For every unlabeled trajectory, OTR orchestrates a series of pivotal steps that effectively intertwine expert behavior with the learning framework.

At the algorithm's core lies the alignment of expert demonstrations with unlabeled trajectories, achieved through the \textsc{SolveOT} function. This pivotal step addresses the OT problem, seeking to establish an optimal coupling between the distributions of expert demonstrations and unlabeled trajectories. During this step, both the cost matrix and the optimal coupling matrix are derived, laying the groundwork for subsequent reward calculation in accordance with Equation \ref{eq:reward_computation}.

Following the successful alignment of distributions, OTR calculates rewards for each state along the trajectory (line 5). The reward computation process is based on the divergence between the expert and the unlabeled trajectory. The calculated rewards are then incorporated into the unlabeled trajectory, effectively "labeling" each state with an associated reward value. This integration of informative rewards transforms an initially unlabeled sequence of states into a reward-annotated dataset for the subsequent RL phase.

Solving the OT problem necessitates addressing a computationally intensive linear program. Given the computational demands, it is crucial to identify an efficient method to tackle this challenge. Different distance metrics, including Gromov-Wasserstein \citep{peyre2016gromov, fickinger2021cross}, Sinkhorn \citep{cuturi2013sinkhorn, papagiannis2022imitation, haldar2023watch}, CO-OT \citep{redko2020co}, GDTW \citep{cohen2021aligning}, and Soft-DTW \citep{cuturi2017soft} have been explored to compute alignments between expert and agent demonstrations. Recent research conducted by Cohen et al. [11] has highlighted Sinkhorn's distance [12] as the most efficient learning method among these metrics. Consequently, OTR employs Sinkhorn's algorithm \citep{cuturi2013sinkhorn} to address an entropy-regularized variant of the OT problem.

A defining characteristic of OTR is its ability to perform offline reward computation. This distinctive feature eliminates the need for real-time sample collection or data manipulation during the RL phase, as rewards are generated prior to the commencement of RL. This attribute enables OTR to integrate reward annotations with a wide range of downstream offline RL algorithms.

\section{Methodology}
In this section, we present our experimental methodology, dataset, and results, aiming to demonstrate the efficacy and versatility of the OTR algorithm in the context of offline RL.

Our experimental framework centers around assessing OTR applicability and effectiveness within a novel domain of offline RL. We investigate whether the OTR algorithm can be seamlessly applied to a new dataset extracted from the ActiveTrack benchmark task within the SurRoL environment. Our objective is to determine if OTR can successfully learn meaningful reward functions and guide policy learning in this distinct domain, offering a unique perspective on its potential applications beyond its original domain.

\begin{figure*}[!t]
\centering
\includegraphics[width=0.99\textwidth]{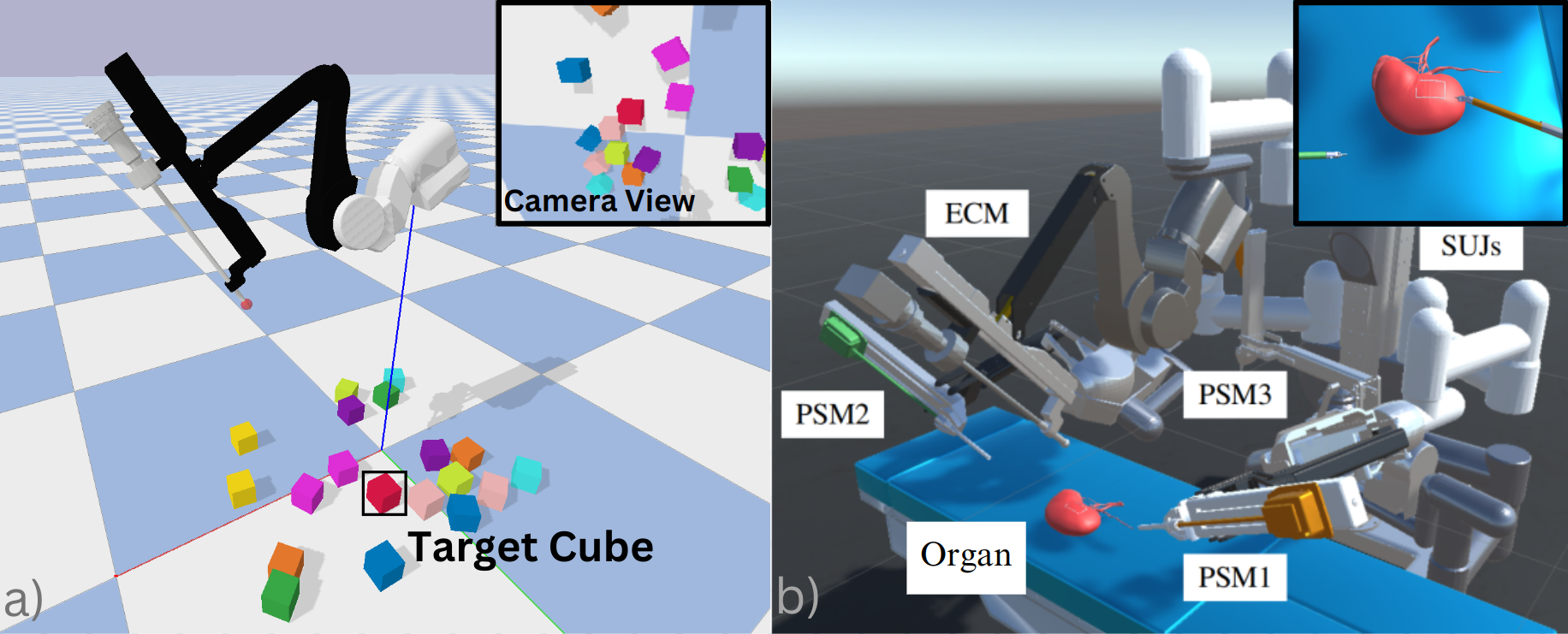}
\caption{\textbf{Surgical Robotics Simulations.} Figure a) showcases the ActiveTrack task, where an ECM, a 4-DoF actuated robot equipped with a camera, is skillfully tracking a moving red cube, simulating the precision required in surgical robotics. The figure presents the ECM, various cubes, including the target cube, and provides a visual representation of the camera view during tracking. Figure b) \citep{fan2022unity} illustrates the application of the ActiveTrack task in real-world surgical robotics scenarios using the da Vinci Surgical Simulator. The setup includes Patient-Side Manipulators (PSMs), an ECM for tracking surgical instruments, Surgical Use Joysticks (SUJs), and an operating table. This configuration showcases how the ActiveTrack task can be employed to precisely follow and track surgical instruments during minimally invasive surgeries.}\label{ecm}
\end{figure*}

Algorithm \ref{otriql} provides a comprehensive overview of a two-step procedure that seamlessly integrates OTR and Implicit Q-Learning (IQL) \citep{kostrikov2021offline}. In the first step, OTR takes expert demonstrations and an unlabeled dataset as its input, utilizing its expertise to calculate an optimal alignment. The result is a labeled dataset enriched with rewards, a crucial component for the subsequent step. In the second phase, we employ IQL as our primary offline RL optimizer to learn a policy. IQL is a versatile Q-learning algorithm that effectively sidesteps the need to query values of actions not present in the dataset during training while still enabling multi-step dynamic programming. This combined approach provides an elegant solution for converting existing offline datasets into a fully trained policy, highlighting OTR's remarkable capacity to render the datasets valuable for policy learning.

\begin{algorithm}[H]
\caption{OTR + IQL}\label{otriql}
\begin{algorithmic}[1]
\Procedure{Step 1: OTR}{}
\State \textbf{Input:} Expert demonstrations \& unlabeled dataset
\State \textbf{Output:} Labeled dataset
\EndProcedure

\Procedure{Step 2: IQL}{}
\State \textbf{Input:} Labeled dataset
\State \textbf{Output:} Trained policy
\EndProcedure

\end{algorithmic}
\end{algorithm}

\subsection{Dataset}
The \textbf{Sur}gical \textbf{Ro}botic \textbf{L}earning (SurRoL) environment is an essential cornerstone of our study, offering a dynamic platform that allows us to explore and evaluate intricate robotic learning scenarios. Designed to replicate the challenges of real-world robotic applications, SurRoL empowers us to simulate complex tasks under controlled conditions, enabling the generation of datasets and the assessment of algorithms in diverse robotic domains.

For this study, we used the ActiveTrack environment from the SurRoL simulation platform, a task meticulously designed to emulate the precision and complexity inherent in surgical robotics (Figure \ref{ecm}). In this setting, an Endoscope Camera Module (ECM) is tasked with tracking a moving red cube in a 2D plane, which acts as a surrogate for a crucial surgical instrument. The action in this environment is the camera's velocity in its own frame and the state encompasses both the robot's pose and the object's pose in the Cartesian space. The environment introduces the challenge of maintaining focus on the primary tool while being perturbed by other cubes that mimic potential distractions during surgery. The red cube follows a square trajectory on a random location within the workspace. The objective is to ensure the ECM's alignment with this moving target, emphasizing surgical precision and dexterity.

Expert demonstrations for this dataset were gathered through a scripted policy designed to emulate the decision-making process of skilled practitioners. By incorporating factors like the cube's position and the ECM's alignment, the policy generates actions reflecting the precision expected from an expert. The policy also calculates camera settings and adjustments based on the cube's position to maintain alignment and refocus the ECM if deviations occur. To obtain the unlabeled dataset, we discard the original reward information in the dataset.

Leveraging these datasets, we proceed to implement the OTR algorithm detailed in the earlier section, adapting it to the specific context of the ActiveTrack environment. The algorithm's efficacy in guiding the ECM to effectively track the moving cube becomes the focal point of our analysis.

\subsection{Implementation}
In our implementation, we have gathered a set of 10 expert demonstrations, each representing distinct and well-defined behaviors within the target environment. Additionally, we have compiled a dataset comprising 100 unlabeled state-action trajectories. Given the presence of multiple expert demonstrations, we independently compute OT for each episode. Subsequently, we select the rewards from the expert trajectory that yields the highest episodic return. This approach ensures that the agent learns from the expert trajectory that offers the most effective guidance, ultimately enhancing the policy learning process. 

The computed rewards undergo a transformation through an exponential function $s(r) = \alpha \exp(\beta r)$. This technique serves the purpose of adjusting the reward values utilized by the offline RL algorithm, aligning them with an appropriate range. Such adjustment is particularly valuable, as the scale of reward values can significantly impact the performance of many offline RL algorithms. In our experiments, we have empirically set $\alpha$ to 5 and $\beta$ to 1 to achieve this adjustment.

To provide robust evaluations and ensure the reliability of our results, we repeat our experiments using 10 different random seeds. During the training phase, the IQL algorithm is run for a total of one million steps. To monitor the RL agent's tracking performance within the ActiveTrack task, we conduct periodic evaluations. Specifically, every 10,000 training steps, the agent undergoes 10 evaluation episodes, and we record the average returns achieved during these evaluations. The hyperparameters and network specifications used in the IQL implementation \citep{kostrikov2021offline} are summarized in Table \ref{hyperpara}.

\newcommand{\specialcell}[2][c]{%
\begin{tabular}[#1]{@{}c@{}}#2\end{tabular}}

\begin{table}[H]
\centering
\caption{\textnormal{\textbf{Hyperparameters.} An overview of hyperparameters used in the IQL algorithm.}}{\label{hyperpara}}
\begin{tabular}{lll}
\toprule
 & \textnormal{Hyperparameter} & \textnormal{Value} \\
\midrule
\multirow{3}{*}{\specialcell{\textnormal{Network}\\\textnormal{Architecture}}} & \textnormal{Hidden layers} & $(256, 256)$ \\
 & \textnormal{Dropout} & None\\
 & \textnormal{Network initialization} & \textnormal{Orthogonal} \\
\midrule
\multirow{3}{*}{\textnormal{IQL}} 
 & Learning rate & $3 \times 10^4$ \\
 & Activation function & \textnormal{ReLU} \\
 & \textnormal{Optimizer} & \textnormal{Adam} \\
 & \textnormal{Discount factor} ($\gamma$) & $0.99$ \\
 & \textnormal{Q-function soft-update rate} ($\tau$) & $0.005$ \\
 & \textnormal{Minibatch size} & $256$ \\
\bottomrule
\end{tabular}
\end{table}

\begin{figure*}[!t]
\centering
\includegraphics[width=.99\textwidth]{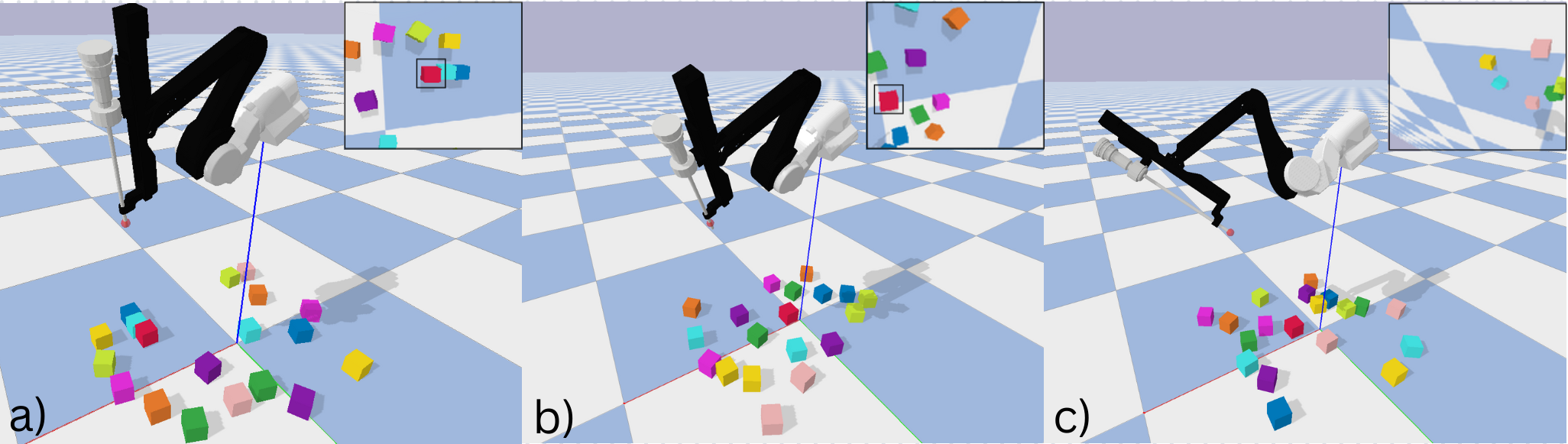}
\caption{\textbf{Visualizing Tracking Scenarios.} The three images depict different tracking scenarios during an ActiveTrack task. In a), the red cube is perfectly centered in the image, signifying maximum tracking precision and yielding the agent a maximum reward of 1. In b), the cube remains within the camera frame but is not precisely centered, resulting in a reward of less than 1. In c), the cube has exited the camera frame, causing a significant drop in the agent's reward. The reward function emphasizes the importance of maintaining focus on the moving cube to maximize tracking efficiency.}\label{t}
\end{figure*}

\begin{figure*}[H]
\centering
\includegraphics[width = .95\linewidth]{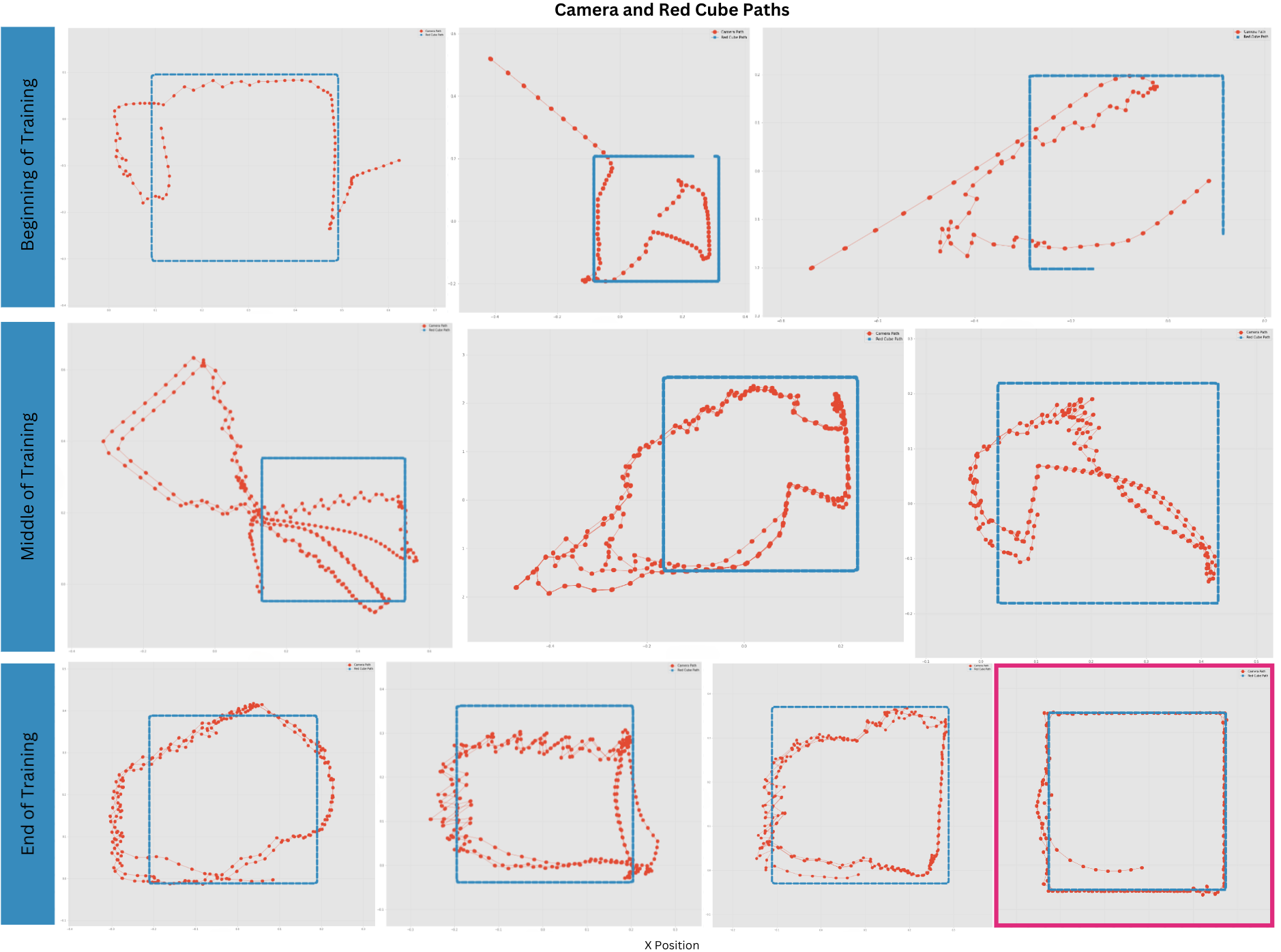}

\caption{\textbf{Tracking Performance Evolution.} The plots in this figure provide insights into the agent's tracking performance at different stages of its training. The circular red marks depicts the points where the center of the camera captures on the plane and the blue square marks track the red cube trajectory. \textbf{First Row (Beginning of Training):}
The first row of plots represents the initial stages of the agent's training. As observed, the camera path struggles to maintain a consistent tracking of the red cube. In some instances, the camera loses sight of the cube even before it can complete a full square trajectory. \textbf{Second Row (Middle of Training):}
Moving to the second row, we examine the agent's performance in the middle of its training. While the camera exhibits improved tracking capabilities, it faces challenges in keeping the red cube consistently at the center of the image. The camera path appears somewhat erratic, indicating occasional deviations from the desired tracking path. \textbf{Third Row (End of Training):}
The third row showcases the agent's performance towards the end of its training. Notably, the camera paths (rightmost three plots) are now able to effectively follow the red cube for the entire episode. Furthermore, these paths exhibit a higher degree of convergence, suggesting that the camera consistently maintains the red cube closer to the center of the image. \textbf{Expert Policy Comparison (Bottom Right):}
The bottom-right plot serves as a reference, demonstrating the tracking proficiency of an expert policy. Comparing the expert's path with the agent's at the end of training reveals significant overlap, suggesting that the agent has successfully learned to emulate the expert's tracking behavior.
}\label{tracking}
\end{figure*}

\subsection{Results}
The ActiveTrack reward function is central to our evaluation of the performance. The reward is computed using the formula:

\begin{equation}
r(s, a) = C - (\|p_t - p_c\|_2 + \lambda \cdot |\theta^*|)
\end{equation}

where $p_t$ signifies the tracked cube's position in image space, $p_c$ stands for the image center, $\lambda$ is a weighting factor, and $\theta^*$ represents the misorientation angle. Misorientation, in this context, refers to the difference between the camera's orientation and the natural line of sight (NLS). Hyperparameters $C$ and $\lambda$ are chosen as 1 and 0.1, respectively.

\begin{figure*}[H]
\centering
\includegraphics[width = .84\linewidth, height = 5.5cm]{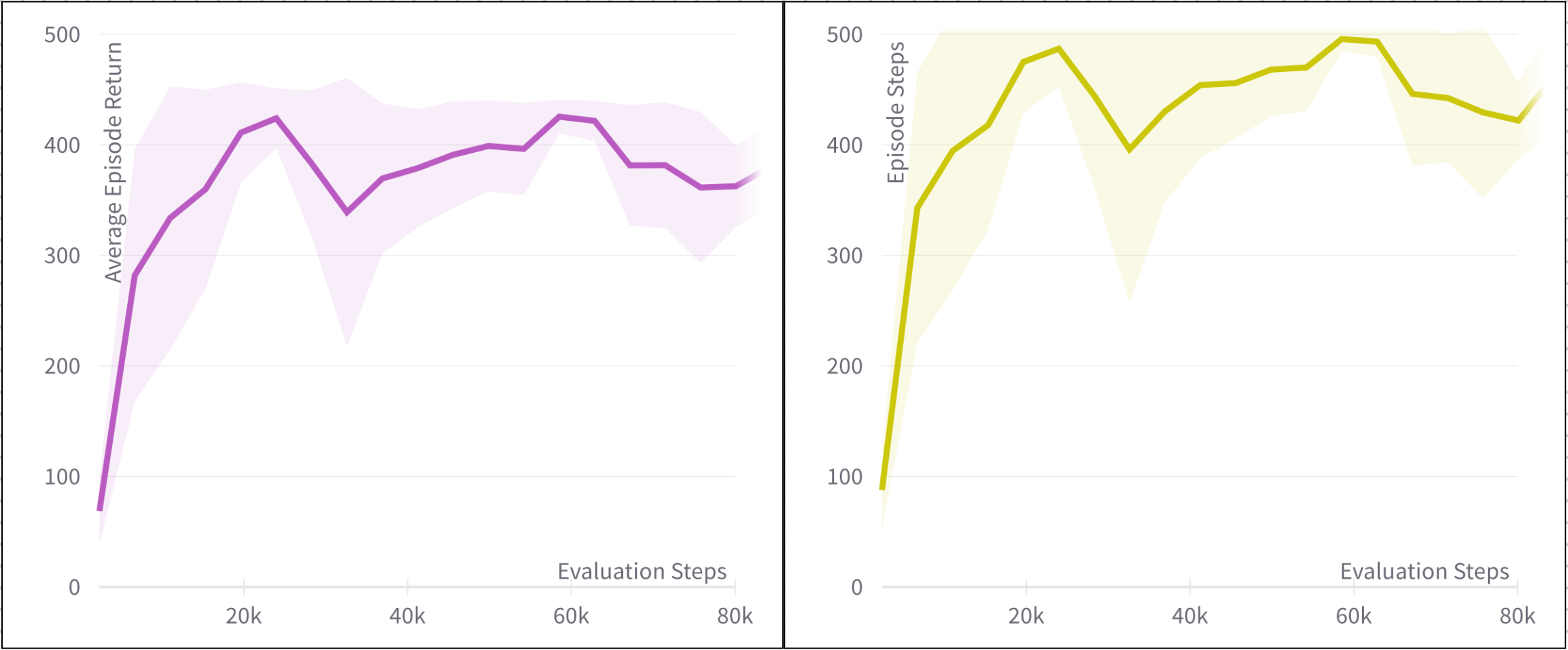}

\caption{\textbf{Experiment Results.} Two line charts present the evaluation results over multiple experiments with 10 random seeds, showing the mean performance with shaded regions indicating $\pm1$ standard deviation across the seeds. The left chart illustrates the total return and the right chart illustrates the episode steps taken during evaluations.}\label{chart}
\end{figure*}

The reward is designed to encourage the agent to maintain focus on the moving cube, ensuring its position in image space ($p_t$) is close to the image center ($p_c$), and its misorientation ($\theta^*$) is minimized. Notably, when the cube is perfectly centered within the image, and the orientation matches the ideal alignment, the agent receives the maximum reward of 1 (Figure \ref{t}). As an episode unfolds, the agent strives to accumulate rewards over a maximum of 500 steps, thus establishing the maximum attainable episode return as 500.

 One of the key metrics of the evaluation was the tracking performance of the RL agent within the ActiveTrack task. In the course of our experiments, we closely monitored the evolution of our model's tracking capabilities as it underwent training. Figure \ref{tracking} illustrates the progressive improvements observed at different stages of the training process. In the initial phases, our agent faced challenges in consistently tracking the red cube. At times, the camera struggled to maintain a reliable view of the cube, often losing sight before it could complete a full square trajectory. As training continued and the model matured, we observed enhanced tracking performance. While occasional deviations persisted, the camera exhibited improved tracking capabilities. Towards the end of training, the model showcased a significant transformation. The camera paths displayed a remarkable ability to effectively follow the red cube for the entire episode. Moreover, these paths demonstrated a higher degree of convergence, indicating that the camera consistently kept the red cube close to the center of the image. To provide context and validation, we also included a comparison with an expert policy in the bottom-right plot, revealing significant overlap with the agent's performance at the end of training.

 To evaluate the tracking performance, we conducted evaluations to gauge how well the agent maintained focus on the moving target throughout its trajectory. Each evaluation consisted of the RL agent tracking the moving red cube for a maximum of 500 steps. Figure \ref{chart} illustrates the results of these evaluations across multiple experiments. The results demonstrate OTR ability to guide the agent effectively, with the agent consistently achieving the maximum episode steps (500 steps) without losing sight of the target. While the agent may not consistently attain the maximum return of 500, it consistently achieves 500 episode steps, demonstrating its ability to maintain the cube within the camera frame even when it's not perfectly centered in the image. This highlights OTR potential to facilitate precise and controlled robotic actions in real-world scenarios. Additionally, the figure illustrates that the agent is capable of achieving high-performance levels relatively early in the training process.

\begin{figure}
\centering
\includegraphics[width = .97\linewidth]{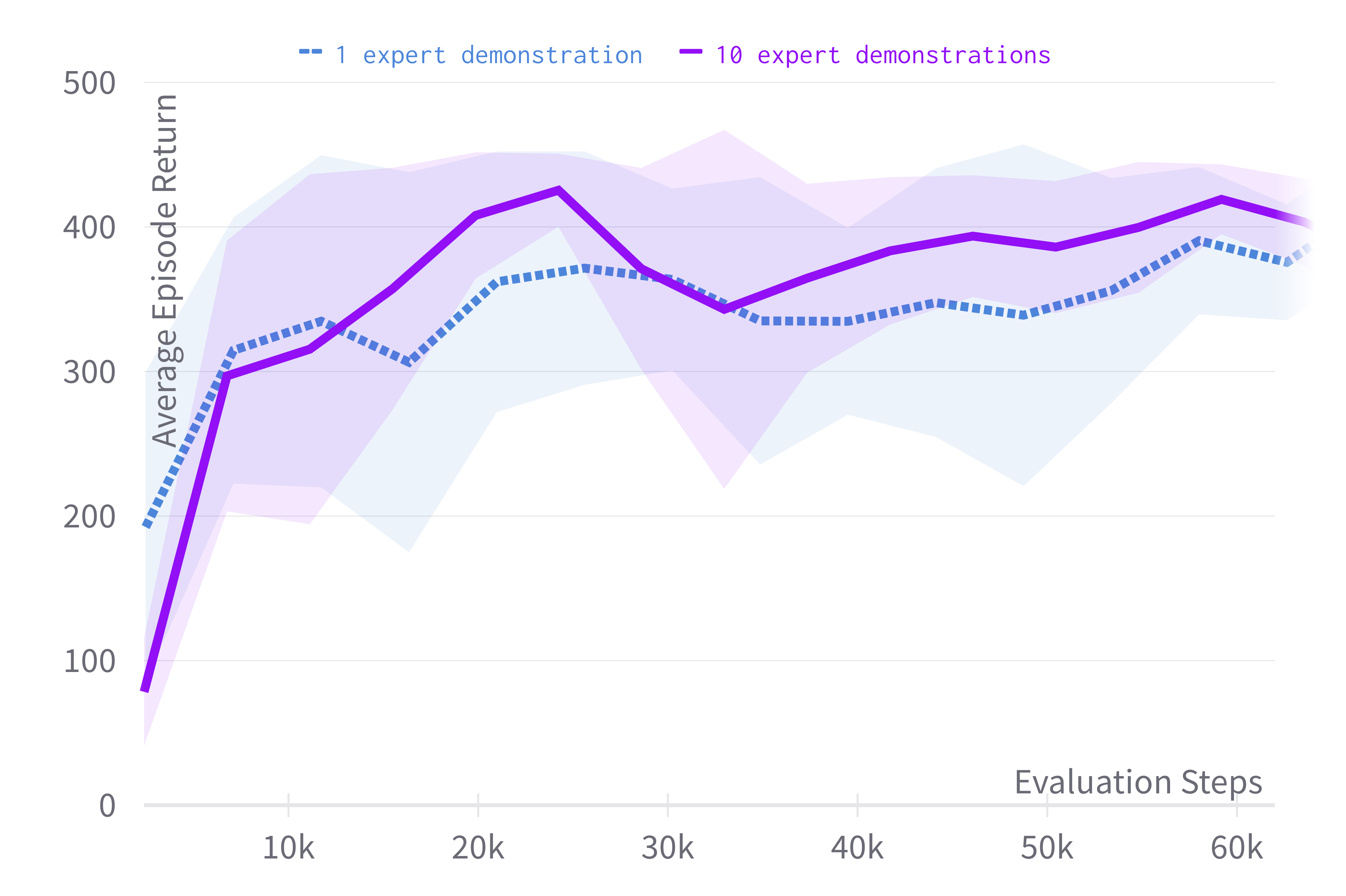}

\caption{\textbf{Performance Comparison Across Different Numbers of Expert Demonstrations.} This figure presents a comparison of the average returns achieved by the OTR-guided RL agent when provided with 10 expert demonstrations (purple) and 1 expert demonstration (blue). The results indicate that while using 10 expert demonstrations results in a slightly higher average return, the difference is relatively small, highlighting OTR's robust performance even with limited expert data.}\label{chart2}
\end{figure}

\textbf{Impact of the number of demonstrations}. Furthermore, we investigated the impact of the number of expert demonstrations on the performance of the OTR-guided RL agent. To this end, we conducted experiments using both 10 expert demonstrations and a reduced set of only 1 expert demonstration. The results, depicted in Figure \ref{chart2}, reveal a noteworthy insight. While employing 10 expert demonstrations does yield a slightly better average return compared to the 1-demonstration scenario, the difference is relatively marginal. This observation underscores the remarkable performance of OTR, even when provided with a limited amount of expert data. It suggests that the algorithm's capacity to autonomously assign rewards based on a single high-quality expert demonstration makes it an efficient and practical choice, reducing the burden of acquiring an extensive set of demonstrations for successful policy learning. This adaptability and robustness in handling a range of expert data quantities further enhance OTR's applicability in real-world scenarios where collecting numerous expert demonstrations may not be feasible. 

 We also evaluate our method by comparing it directly with two state-of-the-art RL algorithms: Soft Actor-Critic (SAC) \citep{haarnoja2018soft} and Deep Deterministic Policy Gradients (DDPG) \citep{lillicrap2015continuous}. We employ the average episode return as the primary evaluation metric, rescaling it linearly within the [0, 1] range for comparative analysis. The performance of each method is presented in Table \ref{compare}. Notably, OTR achieved an average episode return of 0.81, marginally surpassing the figures reported in \citep{huang2023guided}, where SAC achieved 0.79 and DDPG 0.76. OTR employs an extended training duration of 250k steps, in contrast to SAC and DDPG, which use 100k steps. While this longer training period may not conventionally indicate an advantage, it underscores OTR's distinctive approach, which exhibits its own unique characteristics and efficiencies. OTR's approach distinguishes itself by utilizing a fixed dataset of 100 demonstrations, as opposed to the numerous environment interactions typically required. This not only avoids the costs associated with reward formulation and extensive environment interactions but also serves as an implicit investment in improved performance while conserving valuable resources. Importantly, OTR's offline learning paradigm not only adds to its effectiveness but also emphasizes its stability and safety. By removing the need for interacting with the real world during training, OTR mitigates potential safety concerns, eliminating the necessity to deploy partially trained and potentially unsafe policies, and it alleviates the complexities associated with conducting multiple trials or building complex simulators. In conclusion, OTR's extended training period emphasizes its resource-efficient strategy and offline learning advantages, positioning it as a promising option for scenarios prioritizing efficiency, performance, and safety in the learning process.

\begin{table}[h]
\caption{\textbf{Performance Comparison.} Mean scores $\pm$ standard deviations are presented. The results are based on 10 runs with different seeds. OTR was trained for 250k steps, while SAC and DDPG were trained for 100k.}\label{compare}

\resizebox{0.47\textwidth}{!}{
\begin{tabular}{cccc}

\hline
\textrm{
\textbf{Task}} & \textrm{OTR} & \textrm{SAC} & \textrm{DDPG} \\
\hline
 \textrm{ActiveTrack}  & $0.81$ \textcolor{gray}{$\scriptstyle \pm (0.08)$} & $0.79$ \textcolor{gray}{$\scriptstyle \pm (0.08)$}  & $0.67$ \textcolor{gray}{$\scriptstyle \pm (0.08)$} \\
\hline
\end{tabular}
}

\end{table}

\section{Conclusion}
This study introduces the optimal transport reward labelling algorithm and explores its application in the context of offline RL. OTR offers an innovative approach to augmenting unlabeled datasets with reward annotations, which is vital for effective RL agent training. The research demonstrates OTR adaptability by successfully combining it with the IQL algorithm, showcasing its potential to improve policy learning. Furthermore, empirical results highlight OTR efficacy in the context of surgical robotics, particularly within the ActiveTrack task. 

For future directions, this work suggests exploring OTR potentials in various real-world robotic applications and evaluating its scalability and adaptability. Additionally, comparing OTR with other state-of-the-art offline RL and imitation learning methods presents an exciting avenue for future research. Such comparative studies can provide valuable insights into OTR strengths and limitations, paving the way for its adoption across diverse robotic domains.

\section{Acknowledgments}
This research was partially supported by the Australian Research Council’s Discovery Projects funding scheme (project DP190102181 and DP210101465).



\bibliographystyle{cas-model2-names}

\bibliography{paper.bib}

\bio{}
\endbio


\end{document}